\documentclass[twoside]{deon16}

\usepackage{deon16macro}

\usepackage{graphicx}
\usepackage{amsmath}
\usepackage{amssymb}
\usepackage{tikz}
\usepackage{rotating}
\usepackage{pgfplots}

\def\lastname{Schuster}

\begin{document}

\begin{frontmatter}
  \title{Forms and Norms of Indecision in Argumentation Theory}
  \author{Daniela Schuster}\footnote{daniela.schuster@uni-konstanz.de}
 \address{Universität Konstanz \\ 78464 Konstanz}

  \begin{abstract}
One main goal of argumentation theory is to evaluate arguments and to determine whether they should be accepted or rejected. When there is no clear answer, a third option, being undecided, has to be taken into account. Indecision is often not considered explicitly, but rather taken to be a collection of all unclear or troubling cases. However, current philosophy makes a strong point for taking indecision itself to be a proper object of consideration.
This paper aims at revealing parallels between the findings concerning indecision in philosophy and the treatment of indecision in argumentation theory. By investigating what philosophical forms and norms of indecision are involved in argumentation theory, we can improve our understanding of the different uncertain evidential situations in argumentation theory.

  \end{abstract}

 \end{frontmatter}

\section{Introduction}\label{Introduction}

Abstract argumentation theory is concerned with modelling arguments and their relation of defeat. One main goal of argumentation theory is to determine of a given set of arguments which of those should be accepted and which of those should be rejected. This yields an extension -- a set of accepted arguments -- and an antiextension -- a set of of rejected arguments. Additionally, it is possible to distil a third set of arguments that is sometimes not characterised explicitly, which is the set of arguments one is undecided about. The involvement of indecision is made explicit in the labelling-approach of abstract argumentation theory, where arguments are labelled according to three different labels: $\mathrm{in, out ~and~ undecided}$.\\
Very often indecision is only seen as a quite useful but rather unimportant byproduct when characterising the acceptability and rejectability of arguments. From a philosophical point of view, however, besides acceptance and rejection, indecision is the third main doxastic response, and it should be taken to be equally important. Especially in the last couple of years, researchers in the philosophical disciplines of epistemology and philosophy of mind started to focus more on this third, neutral stance. They try to investigate the norms of indecision (which is done in epistemology) and the different phenomena involved in this broad concept (which is done in philosophy of mind), see for example \cite{Wagner.2021} for the distinction.
The philosophical investigations in both of these areas can be very useful when applied to argumentation theory. Investigating the notion of indecision more precisely can help us to observe the different options of how to use indecision as a tool to describe uncertain, doubtful, or conflicting information. This will allow us to find ways to improve the representation of the given knowledge and to make certain decisions less bold and more understandable and trust-worthy.\\
\newline
\noindent
In this paper, I aim at taking the first steps in transferring the mentioned philosophical considerations to argumentation theory and in revealing important parallels. I will shed some light on the different forms of indecision that can be found in abstract argumentation theory and in what way they correspond to particular philosophical phenomena. Moreover, I want to illustrate by what means the various semantics of abstract argumentation theory treat indecision differently and how this relates to the epistemological debate about rationality norms.\\
In a second step, I want to dive deeper into the more structured level of statement-labelling, by considering not only the argument as such, but also its conclusion.
I will attempt to reveal some of the philosophical norms and forms of indecision in these conclusions.
Although statement evaluation has barely been considered explicitly in argumentation theory, this enterprise is even more important from a philosophical perspective, as arguably what determines our actions, is more our beliefs in the form of statements and less the arguments we have in mind themselves.\\
\newline
The paper is structured in the following way.
In Section \ref{Indecision in Philosophy}, some philosophical findings about indecision will be presented, first from the sub-field of epistemology, which is concerned with rationality norms (Section \ref{Norms about Indecision in Philosophy}), and second from the sub-field of philosophy of mind, which can provide insights into the different forms of indecision (Section \ref{Forms of Indecision in Philosophy}). In Section \ref{Indecision in Argumentation Theory} possible transfers from the philosophical findings to the area of argumentation theory will be suggested. In Section \ref{Indecision on the Level of Arguments}, parallels will be drawn between indecision in philosophy and indecision in \emph{abstract argumentation theory}. Here we are operating on the level of arguments. For this, some formal backgrounds will be introduced (Section \ref{Formal Background}), such that the different forms of indecision (Section \ref{Forms of Indecision on the Argument Level}) and the norms of indecision involved in abstract argumentation theory (Section \ref{Norms of Indecision on the Argument Level}) can be revealed. In a second step, in Section \ref{Indecision on the Level of Statements}, parallels between indecision in philosophy and indecision on the statement level of argumentation theory will be considered. Finally, in the outlook (Section \ref{Conclusion and Outlook}), I will conclude these considerations and point to some further areas where the philosophical notion of indecision can be applied fruitfully.
\section{Indecision in Philosophy}\label{Indecision in Philosophy}
In Philosophy, the concept of indecision generally refers to a neutral stance that a person can take towards something. This stance lies somewhere in between believing and disbelieving. Philosophers take the ``something'' one is undecided about (and which is also the object of belief and disbelief)\footnote{For reasons of treating the the three doxastic stances: belief, disbelief and indecision equally, we will assume here that in the philosophical considerations indecision has propositional content, although this has been questioned, for example, in \cite{friedman.2013.question}.} to be a proposition. Although there is quite some dispute about the nature of propositions and their representation, in this paper I will assume that propositions can simply be represented by sentences.
The philosophical investigations concerning indecision can broadly be divided into belonging to one of the two philosophical areas: philosophy of mind and epistemology. In Section \ref{Norms about Indecision in Philosophy}, I will focus on the epistemological part, discussing philosophical norms of indecision. In Section \ref{Forms of Indecision in Philosophy}, the investigations of philosophers of mind will help us to understand different forms of indecision.\\
The terminology concerning indecision is not uniform in philosophy. There are different terms (such as non-belief, indecision, suspension, agnosticism, withholding belief) that are used differently in the literature, see \cite[p. 166]{friedman.2013.suspended}.
I will comment on this briefly in Section \ref{Forms of Indecision in Philosophy}, when describing different forms of indecision. Generally, I will use the term ``indecision'' (in philosophical terminology) to refer to any neutral position that a subject can adopt towards a topic, besides believing or disbelieving it.

\subsection{Norms about Indecision in Philosophy}\label{Norms about Indecision in Philosophy}
Epistemologists focus on providing rules (or norms) for when to adopt a certain doxastic attitude.
Until recently, modern epistemologists were mostly only concerned with formulating norms about when a proposition should be believed or disbelieved, see \cite[p. 165]{friedman.2013.suspended}. The third option of being neutral was not investigated as such, but merely considered a byproduct. At best, the third option of being undecided has been seen as a base case or fall-back position for not-considered propositions before ``proper'' reasoning yields to a decision about the truth of a proposition, i.e., believing or disbelieving it.\\
One can find such philosophical attitudes in the history of philosophy. For example, Descartes formulated in his \emph{Meditationes} \cite{descartes} very strict necessary conditions of when one should believe a proposition. He thought that he should only believe the things he was absolutely certain about. In order to fulfil these very strict rules for belief, he used his approach of methodological doubt and considered every possible statement as undecided.\footnote{In his strategic doubt, Descartes actually first tried to \textit{disbelieve} every proposition he used to belief, so e.g., disbelieve that he has two hands, yielding to his desired point of departure of being in doubt about everything.} In his approach, he laid no restrictions on when something can or cannot be undecided, hence disregarding any norms concerning indecision.\\
\newline
Only recently philosophers became more interested in epistemological concepts tied to this third neutral stance. They started to investigate such phenomena as indecision and formulated norms about when one should or should not be undecided about a proposition.
Friedman’s ignorance norm in \cite{friedman.2017}, for example, suggests a necessary condition for indecision,\footnote{In fact, Friedman is here talking about suspension and not about indecision. As mentioned before, we will use the term ``indecision'' instead and come back to the discussion about the differences in the next sub-section.} stating that a person is only allowed to be undecided about a certain proposition, if she does not know the proposition or its negation already. This requirement forbids being undecided when the proposition in question is already known by the subject. In \cite{friedman.2019} she even suggests a norm that forbids to be undecided on a matter, if one already entertained a belief on that matter. Thereby Friedman formulates a necessary condition for indecision: not knowing or respectively not believing the relevant proposition already. Other norms, in the form of principles, are, for example, discussed in \cite{rosa.2019}.
\noindent
Two very basic norms that one can find in the philosophical literature on indecision are the \emph{Absence of Evidence Norm}, stating that you should be undecided about a proposition $p$, when there is no evidence speaking in favour or against $p$, see \cite[p. 60]{friedman.2013.rational}, and what one could call the \emph{Balanced Evidence Norm}, stating that you should be undecided about $p$, when the evidence for and against $p$ is equally balanced, see for example \cite{schroeder.2012}.\\
In Sections \ref{Norms of Indecision on the Argument Level} and \ref{Indecision on the Level of Statements} I will discuss how these norms about indecision find parallels in argumentation theory. First, however, I want to present some findings from philosophy of mind that show the diversity of the phenomenon. 

\subsection{Forms of Indecision in Philosophy}\label{Forms of Indecision in Philosophy}
Recent investigations in philosophy of mind suggest that there is not just one way in which a person can take a neutral stance or be undecided about a proposition.
\noindent
Most of the philosophers in the current discussion aim at describing what one does if one ``suspends judgement'' or ``suspends belief'' about a matter. Accordingly, these philosophers argue that this is the third \emph{attitude}, besides believing and disbelieving, that one can take towards a proposition. Although current scholars strongly disagree on how to use the term of suspension of judgement, most of them do agree that there are different ways in which a person can be neutral about a proposition. 

\begin{example}\label{example phil}
Let us consider the following example situations of indecision:
\begin{itemize}
\item[1)] A caveman, who is undecided about the proposition that quarks exist. 
\item[2)] Me, being undecided about the proposition that the guava fruit has a lot of vitamin C.
\item[3)] Me, being undecided whether the covid pandemic will be over next year.
\item[4)] Me, being undecided about the proposition that a God exists. 
\end{itemize}

\end{example}
\noindent
Furthermore, let us assume, first, that I only briefly heard about the guava fruit and I never thought about its nutritional values. Secondly, that I do think about the covid pandemic a lot and whether it will be over soon, but that I cannot decide the question by looking at my evidence. Thirdly, let us assume that I also thought about the existence of God, but decided that there is no way to find out, so I will stay undecided.\\
\newline
In the few different situations, different forms of indecision are involved. Situation 1 is a famous example going back to Hájek \cite[p. 205, footnote]{hajek.1998}, that describes a typical case of mere \emph{non-belief}. With the term \emph{non-belief} philosophers describe situations in which a person neither believes nor disbelieves a certain proposition, i.e., a lack of both belief and disbelief. Most philosophers agree that phenomena like suspension of judgement or indecision as such should not be identified with mere non-belief. Wedgwood \cite[p. 272]{wedgwood.2002} for example notices that non-belief is in fact not even an attitude, a position one can take, or a ``mental state'' at all, since, for instance stones can be in this situation of non-belief as well and we would not say that stones can suspend judgement about something. Situation 1 is somehow similar to this, as the caveman could not even understand the concepts involved in the proposition. Clearly, situations 2, 3 and 4 already all involve a more sophisticated form of indecision than situation 1 does.\\
\newline
Even so, philosophy of mind also makes a point for distinguishing situation 2, 3 and 4 from each other.
What we want to describe with suspension of judgement is a \emph{higher} or \emph{more sophisticated} form of indecision; a situation in which we cognitively consider the proposition but are still neutral about its truth. Some sort of ``considering'' or ``thinking about'' condition has to be fulfilled, when we want to say that someone suspends judgement. Although the form of indecision involved in situation 2 already comes closer to suspension of judgement (because I do understand the relevant sentence), I am still lacking the so-called cognitive contact to the proposition, as I have never even thought about it. Most scholars would not describe me as suspending about the proposition that the Guava fruit has a lot of vitamin C, if I never even thought about it.\\
\noindent
This ``considering'' element is given both in situation 3 and 4. In both cases, I had the respective proposition at least temporally in mind. Still, scholars like Friedman \cite{friedman.2013.suspended}, Wagner \cite{Wagner.2021}, or  Raleigh \cite{raleigh.2019} argue that there is still some important and somehow categorical difference between the two situations 3 and 4. In situation 3, although I am undecided, I still continue to think about the question whether the covid pandemic will be over next year. In fact, almost every day I get new evidence speaking for and new evidence speaking against the proposition. Moreover, I might be strongly motivated to find out about the matter, because I might want to plan a trip for next year. This is what distinguishes situation 3 from situation 4, according to \cite{friedman.2013.suspended} or \cite{Wagner.2021}. In situation 4, I ``settled'' the indecision and thereby closed the question about God's existence. Friedman argues that this settlement comes with forming a \textit{sui generis} attitude of suspension, that is not reducible to non-belief in any sense, while Wagner takes the settling element, that is necessary for proper suspension, to be an endorsement of my own indecision. According to this account, non-belief is constitutive but not sufficient for suspension.\footnote{For scholars like Raleigh \cite{raleigh.2019} or Wedgwood \cite{wedgwood.2002} it is sufficient that I form some sort of meta-belief about my own doxastic situation. This can be regarded as a ``committing'' element, too.} Regardless of what the specific accounts take this settling element to be, it can be noticed that in situation 3 this kind of settlement or ``closing the question'' is missing.\\
\newline
To illustrate the differences, I think it can be helpful to set the different forms in relation to each other by visualising them along an axis of ``engaging'' or ``being concerned'' or ``being involved'' with a proposition.

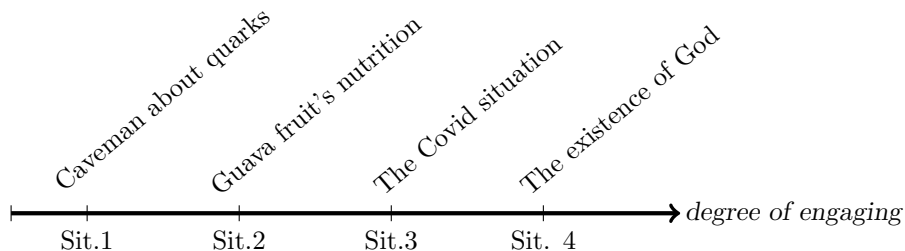
\begin{center}
    \begin{tikzpicture}
        \draw[ultra thick,->] (-5,0) -- (3.8,0);

    \foreach \x in {-5,-4,-2,0,2}
    \draw (\x cm,3pt) -- (\x cm,-3pt);
    
    \draw (-5,0) node[below=3pt] {} node[above=3pt] {};
    \draw (-4,0) node[below=3pt] {Sit.$1$} node[above=3pt] {};
    \draw (-3,0) node[below=3pt] {} node[above=3pt] {$\begin{turn}{40} 
        Caveman about quarks
        \end{turn}$};
    \draw (-2,0) node[below=3pt] {Sit.$2$} node[above=3pt] {};
    \draw (-1,0) node[below=3pt] {} node[above=3pt] {$\begin{turn}{40} Guava fruit's nutrition \end{turn}$};
    \draw (0,0) node[below=3pt] {Sit.$3$} node[above=3pt] {};
    \draw (1,0) node[below=3pt] {} node[above=3pt] {$\begin{turn}{40}  The Covid situation \end{turn}$};
    \draw (2,0) node[below=3pt] {Sit. $4$} node[above=3pt] {};
    \draw (3,0) node[below=3pt] {} node[above=3pt] {$\begin{turn}{40} The existence of God  \end{turn}$};
    \draw (5.3,0) node {\textsl{degree of engaging}} node[above=3pt] {};
\end{tikzpicture}
\end{center}

\noindent
The caveman does not engage with the proposition about the existence of quarks at all, as he does not even understand it. While I do understand the proposition about the nutrition of the Guava fruit, I still do not really engage with it, as I never even entertained it in thought. Compared to this, I engage with the proposition about the covid pandemic a lot. In fact, I constantly think about it. Situation 4, though, falls a bit out of the pattern. One the one hand one could argue that I, in fact, engage less with the proposition about God's existence than with the proposition about the Covid pandemic, in the sense that I do not think about the proposition constantly anymore. On the other hand, however, I am still in some sense more \emph{bound} to the proposition in situation 4, as I definitely include the proposition into (the third category of) my belief sets. In this sense, although I do not think about it a lot anymore, I am yet more identified and involved with the proposition and I am further in the evaluating process than I am with the proposition about the covid pandemic. I closed the question about God's existence and I integrated the proposition into my doxastic household. In this sense situation 4 has to be placed on the very right of the axis.\\
\newline
In the next section, I will relate these different philosophical forms and the epistemological considerations from Section \ref{Norms about Indecision in Philosophy} to the field of argumentation theory.

\section{Indecision in Argumentation Theory}\label{Indecision in Argumentation Theory}
As stated in the introduction, the main goal of this paper is to connect the usage of indecision in argumentation theory with the philosophical concept of indecision. We will do this on two levels:  first on the argument level (Section \ref{Indecision on the Level of Arguments}) and second on the statement level (Section \ref{Indecision on the Level of Statements}). The doxastic trisection of philosophy (believing, disbelieving and indecision) finds correspondence both on the argument level of abstract argumentation theory and on the statement level of structured argumentation theory. As described above, philosophers take propositions, that are structurally equal to statements or sentences, to be the object of the doxastic stances. Hence, parallels between philosophy and the statement level of argumentation theory are easier to draw than the parallels with the argument level. At the argument level, in abstract argumentation theory, arguments are the relevant objects for evaluation, which are structurally very different from propositions. Nevertheless, a correspondence between the three doxastic responses one can take towards a proposition in philosophy and three possible reactions one can show towards an argument in abstract argumentation theory can still be found.

\subsection{Indecision on the Level of Arguments}\label{Indecision on the Level of Arguments}
\subsubsection{Formal Background}\label{Formal Background}
Before I can shed some light on the parallels between indecision in abstract argumentation theory and philosophy, some basic notions of abstract argumentation theory have to be introduced. This sub-section does not aim at providing a complete overview of the definitions and theorems of abstract argumentation theory. Only the parts that are directly relevant for the considerations concerning indecision are presented. The following definitions can for example be found in \cite{baroni.2011}.\\
Abstract Argumentation Theory focuses on modelling arguments and their relation of attack on an abstract level. This is modelled in argumentation frameworks, which allow for an illustration of arguments (nodes of the graph) and their attack (edges of the graph). More precise features, such as the internal structure of the arguments and how they attack each other is not representable at this abstract level.
\begin{definition}[Argumentation Framework]
An Abstract Argumentation Framework is a pair $(Ar, att)$, where $Ar$ is a set of arguments and $att\subset Ar \times Ar$ is a relation of attack between the arguments.
\end{definition}
\noindent
Besides the modelling challenge, abstract argumentation theory also deals with evaluating arguments in order to choose a proper subset of acceptable arguments among the modelled ones.\\
In general, there are two approaches for this: the extension approach and the labelling approach.
The labelling approach provides a function that maps each argument to a label, that is either $\mathrm{in}$, $\mathrm{out}$ or $\mathrm{undec}$ (undecided).

\begin{definition}[Labelling]
Given a set of labels, $\{\mathrm{in, out, undec}\}$, a labelling is a function $Lab: Ar \rightarrow \{\mathrm{in, out, undec}\}$ which maps each argument to one of the three possible labels.
\end{definition}
\noindent
In contrast to this, there is the extension approach. The extension approach simply yields a subset of the considered arguments (the extension) which consists of the accepted arguments. The arguments that are rejected or undecided are then defined in terms of the extension.
Although the two approaches are convertible into each other, see \cite{baroni.2011}, the labelling approach provides a straight-forward way to distinguish the three possible states of an argument and speaks about an undecided evaluation explicitly. Therefore, I will focus on the labelling approach in this paper.\\
\newline
The challenge then is to decide which argument should get which label. This decision is not to be made arbitrarily, but has to follow certain rules. These rules are given by different so-called semantics.
Quite basic rules are provided by admissible semantics.

\begin{definition}[Admissible Semantics]
Let $(Ar, att)$ be an argumentation framework. A labelling $Lab$ is called an admissible labelling (or a labelling according to admissible semantics), iff the following two conditions hold:
\begin{itemize}
    \item every $\mathrm{in}$-labelled argument $A\in Ar$ is legally $\mathrm{in}$, i.e., $\forall B \in Ar$: if $(B,A)\in att$ then $Lab(B)=\mathrm{out}$.
    \item every $\mathrm{out}$-labelled argument $A\in Ar$ is legally $\mathrm{out}$, i.e., $\exists B \in Ar$, such that $(B,A)\in att$ and $Lab(B)=\mathrm{in}$.
\end{itemize}
\end{definition}    
\noindent
Admissible semantics demand that only arguments that are \textit{legally} $\mathrm{in}$ or $\mathrm{out}$, should get the respective label. Informally speaking, an argument is legally $\mathrm{in}$, if all its attackers (if there even are any) are labelled $\mathrm{out}$. An argument is legally $\mathrm{out}$, if it has at least one attacker that is labelled $\mathrm{in}$.\\
\newline
Another semantics, that we we will introduce here and need in Section \ref{Norms of Indecision on the Argument Level}, as it differs from admissible semantics in some important manner, is complete semantics. 
\begin{definition}[Complete Semantics]
Let $(Ar, att)$ be an argumentation framework. A labelling $Lab$ is called a complete labelling (or a labelling according to complete semantics), iff the following three conditions hold:
\begin{itemize}
    \item every $\mathrm{in}$-labelled argument $A\in Ar$ is legally $\mathrm{in}$, i.e., $\forall B \in Ar$: if $(B,A)\in att$ then $Lab(B)=\mathrm{out}$.
    \item every $\mathrm{out}$-labelled argument $A\in Ar$ is legally $\mathrm{out}$, i.e., $\exists B \in Ar$, such that $(B,A)\in att$ and $Lab(B)=\mathrm{in}$.
    \item every $\mathrm{undec}$-labelled argument $A\in Ar$ is legally $\mathrm{undec}$, i.e., $ \exists B \in Ar$, such that $(B,A)\in att$ and $Lab(B)\not=\mathrm{out}$ and $\forall B \in Ar$: if $(B,A)\in att$ then $Lab(B)\not=\mathrm{in}$
\end{itemize}
\end{definition}
\noindent
Complete semantics fulfil the requirements of admissible semantics and fulfil on top of that the requirement that every $\mathrm{undec}$-labelled argument has to be also legally $\mathrm{undec}$. An argument is legally $\mathrm{undec}$, iff it is neither legally $\mathrm{in}$ nor legally $\mathrm{out}$. This means that an argument is legally $\mathrm{undec}$ iff \emph{none} of its attackers are labelled $\mathrm{in}$ and it has at least one attacker that is \emph{not} labelled $\mathrm{out}$.\\
\newline
Admissible and complete semantics are only two of the many possible semantics that have been investigated in argumentation theory. All semantics make demands on the labellings of the arguments. Some  semantics (e.g., grounded semantics) only yield one allowed labelling. Most semantics, like admissible and complete semantics, however, usually allow for a plurality of possible labellings. In such situations the question of the justification status of an argument is raised. If an argument is, for example, labelled $\mathrm{in}$ according to one complete labelling and labelled $\mathrm{out}$ in another one, how should we evaluate that argument after all?\\
A rather detailed analysis of different justification statuses is provided in \cite{wu.2010}. Wu et. al. consider in particular the case of complete semantics and note that the labellings produced by complete semantics can be interpreted as the different reasonable stances one can take towards an argument. 
They suggest a function that maps each argument to its justification status, which they take to be the set of all possible labels each argument gets from the different complete labellings.
\begin{definition}[Justification Status]
Let $(Ar, att)$ be an argumentation framework. For $A\in Ar$ let $J(A)$ be the justification status of $A$, given by the function $J: Ar \rightarrow \mathcal{P}(\{\mathrm{in, out, undec}\})$.\cite[p. 16]{wu.2010}
\end{definition}
\noindent
$J$ maps every $A\in Ar$ to a subset of $\{\mathrm{in, out, undec}\}$ consisting of all the labels $A$ gets by one or more complete labelling.\\
Considering complete semantics, in \cite[p. 16]{wu.2010} 6 possible justification statuses are obtained:\footnote{Note that the two remaining options $\{\emptyset \}$ and $\{\mathrm{in, out}\}$ are not considered, as \cite{wu.2010} only take non-empty labelling-sets into account and complete semantics have the characteristic of being ``abstention allowing'', meaning that if there is a complete labelling that labels $A$ $\mathrm{in}$ and there is another complete labelling that labels $A$ $\mathrm{out}$, there has to be a third one that labels $A$ $\mathrm{undec}$, \cite[p. 27]{baroni.2011}.}
$\{\mathrm{in, out, undec}\}, \{\mathrm{in, undec}\}, \{\mathrm{out, undec}\}, \{\mathrm{in}\}, \{\mathrm{out}\}, \{\mathrm{undec}\}.$\\
\newline
The basic idea behind justification statuses and about abstract argumentation theory in general can be visualised in the following example:

\begin{example}\label{example AT}
Imagine you get introduced to a group of friends and you are trying to evaluate the different people and their relationships. You get the following information:\footnote{The example is an adapted version from the example in \cite[p. 21]{wu.2010}.}\\
$A$: Alice says that Carole is a Liar.\\
$C$: Carole says that David is a Liar.\\
$D$: David says that Alice is a Liar.\\
$B$: Everybody agrees that Bob is really trustworthy.\\
$E$: Emily says that Fred is a funny guy.\\
$F$: Fred says that Emily does not know him.\\
\newline
\noindent
The argumentation graph of the argument can be visualised like this:\\

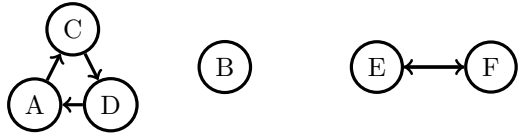
\begin{center}
\begin{tikzpicture}
    \node[shape=circle,draw=black,very thick] (A) at (5,0) {A};
    \node[shape=circle,draw=black,very thick] (B) at (7.5,0.5) {B};
    \node[shape=circle,draw=black,very thick] (C) at (5.5,1) {C};
    \node[shape=circle,draw=black,very thick] (D) at (6,0) {D};
    \node[shape=circle,draw=black,very thick] (E) at (9.5,0.5) {E};
    \node[shape=circle,draw=black,very thick] (F) at (11,0.5) {F} ;

    \path [very thick,->] (A) edge node[left] {} (C);
    \path [very thick,->](C) edge node[left] {} (D);
    \path [very thick,->](D) edge node[left] {} (A);
    \path [very thick,->](E) edge node[left] {} (F);
    \path [very thick,->](F) edge node[right] {} (E);

\end{tikzpicture}
\end{center}

\noindent
The admissible labellings of the given graph are the following:\\

\[
\begin{array}{lll}
\bigl\{\{B,F\},\{E\},\{A,C,D\}\bigr\}, &
\bigl\{\{B,E\},\{F\},\{A,C,D\}\bigr\}, & \bigl\{\{B\},\emptyset,\{A,C,D,E,F\}\bigr\}, \\
\bigl\{\{F\},\{E\},\{B,A,C,D\}\bigr\}, &
\bigl\{\{E\},\{F\},\{B,A,C,D\}\bigr\}, &
\bigl\{\emptyset,\emptyset,\{B,A,C,D,E,F\}\bigr\},
\end{array}
\]
while only the first three are also complete labellings.\footnote{Note that the notation $\bigl\{\{A\},\{B\},\{C\}\bigr\}$ corresponds to a labelling in which $A$ is labelled $\mathrm{in}$, $B$ is $\mathrm{out}$ and $C$ is $\mathrm{undec}$.}\\

\noindent
The justification status of the arguments (according to complete semantics) are:\\
\[
\begin{array}{lll}
   J(A)=\{\mathrm{undec}\},  & J(C)=\{\mathrm{undec}\}, & J(D)=\{\mathrm{undec}\}, \\
  J(B)=\{\mathrm{in}\}, &
J(E)=\{\mathrm{in, out, undec}\}, &
J(F)=\{\mathrm{in, out, undec}\}.
\end{array}
\]
\end{example}

\subsubsection{Forms of Indecision on the Argument Level}\label{Forms of Indecision on the Argument Level}
As we have seen, in abstract argumentation theory arguments are generally evaluated by labelling them according to the three labels: $\mathrm{in, out, undec}$. Although philosophy is concerned with propositions rather than arguments, this trisection still finds some correspondence. Philosophers are concerned with determining when a person does (or should) believe, disbelieve or be undecided about a proposition. Believing a proposition can be said to correspond to labelling an argument $\mathrm{in}$, disbelieving to labelling it $\mathrm{out}$ and being undecided to labelling it $\mathrm{undec}$. If a subject believes a proposition, she takes this proposition to be true, if we label an argument $\mathrm{in}$, we accept it to be valid and possibly even sound and thereby take its conclusion (and its premises) to be true.\\
Within this correspondence, parallels between the different forms of indecision in philosophy and the different forms of indecision in argumentation theory can be drawn. These parallels can be useful to transfer important insights from philosophy to argumentation theory and vice versa. The parallels can then be used to also apply further developments in the philosophical considerations about indecision to argumentation theory.\\
\newline
The authors in \cite{wu.2010} interpret arguments with justification status $\{\mathrm{out}\}$ to be clearly (or strongly) rejected and arguments with the justification status $\{\mathrm{in}\}$ to be strongly accepted. In comparison to that, they interpret arguments with justification status $\{\mathrm{in, undec}\}$ to be only weakly accepted (and arguments with justification status $\{\mathrm{out, undec}\}$ respectively only weakly rejected). So, two of the six possible statuses represent (two forms of) acceptance and another two represent rejection. Finally, the remaining two justification statuses $\{\mathrm{in, out, undec}\}$ and $\{\mathrm{undec}\}$ represent indecision. In \cite{wu.2010} $\{\mathrm{in, out, undec}\}$ is called an undetermined borderline case, while $\{\mathrm{undec}\}$ is taken to be a determined borderline case.\\
\newline
The difference between determined and undetermined resides exactly in the difference between committed and uncommitted indecision. In the case of the justification status $\{\mathrm{in, out, undec}\}$ all options are still available, i.e., the argument may be labelled $\mathrm{in}$, it may be labelled $\mathrm{out}$ and it may be labelled $\mathrm{undec}$. In the determined borderline case of the justification status $\{\mathrm{undec}\}$, it is decided that the argument has to be labelled $\mathrm{undec}$. In all possible (in this case complete) labellings the argument is labelled $\mathrm{undec}$. Hence, the question ``What label does the argument get?'' is no longer open. The only plausible label in this case is $\mathrm{undec}$.
In comparison to this, in the undetermined borderline case of the justification status $\{\mathrm{in, out, undec}\}$, the different labellings disagree on how to label the argument. This can be interpreted as a situation of a vote for which there are some voices voting for $\mathrm{in}$, some for $\mathrm{out}$ and some for $\mathrm{undec}$. It is also possible to interpret this situation within one reasoning subject. The different voters would then correspond to different points of evidence \emph{one} subject has, that are pointing in different directions, i.e., towards $\mathrm{in}$, $\mathrm{out}$ or $\mathrm{undec}$. Both interpretations take the question of what label the argument should get to be not settled, i.e., to be still an open question.
This shows us that the difference between suspension of judgement as a settled form of indecision and other non-committing forms of ``mere indecision'', investigated by philosophers, is reflected in the two different undecided justification statuses an argument can get. In the example situations from section \ref{Forms of Indecision in Philosophy}, we would, give the proposition\footnote{Again, note the different object of the justification statueses: arguments in the case of abstract argumentation theory vs. propositions in the philosophical example.} that the covid pandemic will be over next year, the justification status $\{\mathrm{in, out, undec}\}$, as it is still an open question whether we will accept, reject or stay undecided about the matter, whereas the proposition that a God exists will get the committing justification status $\{\mathrm{undec}\}$, as it is a closed question in the sense that I committed myself to be undecided about it. \\
\newline
This can also be visualised in Example \ref{example AT}. While argument $A$, $C$, and $D$ get the justification status $\{\mathrm{undec}\}$, argument $F$ and $E$ get the justification status $\{\mathrm{in, out, undec}\}$. In the case of $A$, $C$ and $D$ each argument attacks one of the other three, yielding a circle of attack.
Given the information of the example, there is no way to decide who is a Liar and who is not.\footnote{Of course this is a reformulation of the famous Liar Paradox.} Hence, the only reasonable thing to do is to label all three arguments $\mathrm{undec}$. The justification status expresses this commitment to the indecision. On the other hand, for the arguments $F$ and $E$ we have more options. We could label both $\mathrm{undec}$, or we could, for example, label $F$ $\mathrm{in}$ and $E$ $\mathrm{out}$. In this scenario, we might believe that Fred has a good insight on who of the group actually knows him and, therefore, we should believe him when he says that Emily does not know him. Then, however, we should not trust any of Emily's statements about Fred, and argument $E$ is labelled $\mathrm{out}$.
The situation for argument $E$ and $F$ is more open and flexible, and less committed to indecision, than the situation for $A$, $C$ and $D$.\\
\newline
One can observe that the different justification statuses of $E$ and $F$ compared to $A$, $C$ and $D$ do not stem from the different arguments as such, but simply from the fact that $E$ and $F$ are involved in an \emph{even} cycle of attack, while $A$, $C$ and $D$ form an \emph{odd} cycle of attack. This can be seen more clearly when we change the argument $E$ to ``Emily says that Fred is a Liar'' and $F$ to ``Fred says that Emily is a Liar''. Although, in terms of content, we seem to have the same situation with $E$ and $F$ now as we have with $A$, $C$ and $D$, the arguments $E$ and $F$ will still get the justification status $\{\mathrm{in, out, undec}\}$.\footnote{I thank an anonymous reviewer for stressing this point.} 
The situation of $A$, $C$ and $D$, i.e., the situation that all involved arguments \emph{have to} be labelled $\mathrm{undec}$ in fact only occurs in odd-length attacking cycles. This was observed in \cite[p. 242]{pollock.2001} and also revisited later, e.g., in \cite[p. 21]{baroni.2011}. 
Many scholars regard the unequal treatment of odd and even cycles in semantics like complete semantics as problematic. Therefore, other not-admissible based semantics have been developed that allow for an equal treatment of the cycles, see \cite{baroni.2005}.\\
For the purpose of this paper, however, it is only relevant that there are cases, in which the justification status is purely $\{\mathrm{undec}\}$ and that those cases can be seen to represent a committed form of indecision, as in the example above and that there also are cases of arguments with the non-committing justification status $\{\mathrm{in, out, undec}\}$.\\
Although I think that the example as it is presented in \ref{example AT} is better suited than the adapted example to describe, why the situation for $E$ and $F$ is less committed, than the situation for $A$, $C$ and $D$, there is still some difference between the situations, even when changing the content of argument $E$ and $F$ to the respective Liar sentences. The difference consists in how ``easy'' it is for a person that is presented with the arguments to give the arguments a label that is different from $undec$. Although in the adapted example of $E$ and $F$, it still might seem to be the most reasonable action to label both arguments $undec$, you can still do otherwise. If you, for example, happen to believe women more than men, you might think that Emily is more trustworthy. In believing what Emily says, you thereby believe that Fred is a Liar and take argument $E$ to defeat argument $F$. Then you will reject argument $F$ and thereby all of the attackers of $E$ are labelled $out$, allowing you to label $E$ $in$. This is not possible in the same way for the threefold attacking cycle of $A$, $C$ and $D$. Say, that you, for some external reason, believe Alice. Hence, you will believe what she says and argument $A$ will successfully defeat argument $C$. You take Alice' word and believe that Carole is a Liar and hence you will reject argument $C$. Then, argument $D$ has no (not-rejected) attacker anymore; so that you should accept it. As argument $D$, however, tells you that Alice is a Liar, you will have to question your starting point, which is that you belief Alice, again. One can argue that in such odd cycles of attack, you will not manage to break out of the cycle and hence you will not be able to react otherwise than to take each argument to be undecided.\\
\newline
In the next section, I will move from considerations about different forms of indecision in abstract argumentation theory to the involved norms concerning indecision.

\subsubsection{Norms of Indecision on the Argument Level}\label{Norms of Indecision on the Argument Level}
In Section \ref{Norms about Indecision in Philosophy}, I described the development in epistemology that started with not considering explicit norms for indecision and treating indecision merely as a lack of belief and disbelief and as a fallback position for propositions that do not fulfil the requirements of belief and disbelief or are simply not investigated yet. This was typified by the story of Descartes. Only in the last couple of years, epistemologists started to investigate norms about indecision specifically. This development finds some correspondence in the different semantics of argumentation theory.\\
\newline
One can note that admissible semantics only formulate rules (or necessary conditions) about when an argument can be accepted (labelled $\mathrm{in}$) or rejected (labelled $\mathrm{out}$). There is no such requirement for when an argument is labelled $\mathrm{undec}$. A labelling that simply labels all arguments $\mathrm{undec}$ always fulfils the requirements of admissible semantics.
This means that admissible semantics only provide \textit{norms} (in the form of restrictions) for acceptance and rejection, but no norms about when an argument should or should not be regarded as undecided. Hence, the label $\mathrm{undec}$ is always suitable for any argument. \\
Complete semantics, on the other hand, expand admissible semantics (as they adopt the necessary conditions on when an argument is legally $\mathrm{in}$ or $\mathrm{out}$), but also provide restrictions on when an argument can be labelled $\mathrm{undec}$. The treatment of the undecided-label is what distinguishes complete semantics from admissible semantics. While in admissible semantics, the set of undecided arguments is basically only a collection of whatever is left over, (because it has not been determined yet or because it does not manage to be labelled $\mathrm{in}$ or $\mathrm{out}$,) in complete semantics there are rules about when an argument can legally be called undecided.\footnote{The attitude that the $\mathrm{undec}$ label represents arguments for which there is no decision yet, although there should be a clear decision is even better illustrated in preferred or stable semantics, that try to maximise the number of accepted and rejected arguments or even try to bring the set of undecided arguments to be the empty set, see \cite{baroni.2011} for an overview}\\ The difference is also visible in Example \ref{example AT}. While the last labelling, in which every argument is labelled $\mathrm{undec}$, is admissible, this labelling is not complete. In particular, complete labellings forbid to label the argument $B$ that is not attacked by any other argument $\mathrm{undec}$. 
This rule can be interpreted in philosophical terminology as a norm stating that, if there is good evidence for a proposition, then one should not be undecided about it, but believe it, and if there is good evidence against a proposition, then one should not be undecided either, but disbelieve it.\\
\newline
\noindent
Moreover, the different justification statuses also show some parallels with the different norms of indecision from epistemology. In \cite[p. 17]{wu.2010}, it is proven that an argument $A$ gets the justification status $\{\mathrm{undec}\}$, iff $A$ is not labelled $\mathrm{in}$ in any admissible set \emph{and} $A$ is not labelled $\mathrm{out}$ in any admissible set.
Moreover, an argument $A$ gets the justification status $\{\mathrm{in, out, undec}\}$, iff $A$ is labelled $\mathrm{in}$ in at least one admissible labelling \emph{and} $A$ is labelled $\mathrm{out}$ in at least one admissible labelling, see \cite[p. 19-20]{wu.2010}.
The first case can be interpreted as having no evidence for or against the argument (or having no one voting pro or con the argument), while the second case can be interpreted as having evidence for and evidence against the argument (or having both voices that vote for the argument and against it). Hence, the first case corresponds to the absence of evidence norm from philosophy and the second case to the balanced evidence norm, that were introduced in Section \ref{Norms about Indecision in Philosophy}.\\
It can be concluded that different philosophical norms concerning indecision as well as the epistemological development are represented on the argument level of argumentation theory. Next, I want to consider statements, instead of arguments and reveal parallels with philosophical investigations on this level, too.

\subsection{Indecision on the Level of Statements}\label{Indecision on the Level of Statements}
Up to now, I have only addressed the evaluation of arguments as such. It can be argued, however, that it is not so much the arguments themselves, but sentences or propositions that form the doxastic situation of an agent, on which the agent ultimately bases her actions on. In order to reveal this doxastic situation, one has to dive deeper and extract the inner structure of the arguments, i.e., revealing the premises and conclusions involved. By doing this, we end up on the level of statements.
In the following sub-section, I will present some transfers of the philosophical findings concerning forms and norms of indecision to the statement level of argumentation theory.
When philosophical considerations come into play, the concentration on the statement level is particularly desirable, because statements are structurally similar to propositions, which are, as described, the main objects of investigation in philosophy. The parallels can therefore be drawn more straight-forwardly.\\
\newline
Considering all this, it is quite astonishing that assessing justification statuses to statements has gotten, in comparison to arguments, only little attention. Just recently, Baroni and Riveret describe in \cite{baroni.2019} possibilities to transfer the justification of arguments to justification of statements by evaluating the conclusions of the arguments in question.\footnote{In fact Wu et. al. consider in \cite{wu.2010} also briefly how to transfer the justification statuses to statements.} In general, they distinguish two approaches, which they call an argument-focused and a statement-focused approach, to determine the status of statements. In both approaches the status of a statement is eventually determined by the statuses of the arguments speaking for or against that statement. For the purpose of this paper, the different approaches are not directly relevant. What I will focus on here, is the different types of statement labellings that are considered in \cite{baroni.2019}. They distinguish basically between three types of statement labellings.
\begin{itemize}
    \item Bivalent Labellings: Bivalent statement labellings only allow for two possible labels a statement can obtain. \cite[p. 839]{baroni.2019} take the possible labels to be $\{\mathrm{yes, no}\}$.
    \item Doubt-Tolerant Labellings: doubt-tolerant labellings allow for a third label between the definite answers of $\mathrm{yes}$ and $\mathrm{no}$ of the bivalent labellings. This third, intermediate label expresses some sort of doubt. Baroni et. al. call in \cite[p. 848]{baroni.2019} the three labels $\{\mathrm{yes, fal, ni}\}$ where $\mathrm{yes}$ means that a statement is accepted (or verified), $\mathrm{fal}$ means that a statement is falsified and $\mathrm{ni}$ means that there is doubt about the status of the statement.
    \item Ignorance-Aware Labllings: ignorance-aware labellings further divide the group of undetermined statements. Besides $\mathrm{yes}$ and $\mathrm{fal}$ there are two intermediate labels: $\mathrm{unk}$ and $\mathrm{ni}$, yielding the set $\{\mathrm{yes, fal, unk, ni}\}$, where $\mathrm{unk}$ stands for unknown statements and is meant to capture statements for which there is no evidence or lack of knowledge, and $\mathrm{ni}$ captures the statements for which the evidence indicates indecision. 
\end{itemize}
The differences of the three labelling-types can be understood best, when one takes a look at the following example from \cite{baroni.2019}, which was introduced in \cite[p. 489]{baroni.2016}:\\
``Suppose that Dr. Smith says to you: ‘Given your clinical data I conclude you are
affected by disease D1’. Suppose then that another equally competent physician Dr. Jones says to
you: ‘Given your clinical data I conclude you are not affected by disease D1’. Your view on the
justification of the statements s1=‘I am affected by disease D1’ and $\lnot$s1=‘I am not affected by
disease D1’ may become quite uncertain. In a different situation, at home, you use an off-the-shelf test kit suggesting you have caught
disease D2. You then undertake a serious and reliable clinical test, which excludes disease D2.
Would you consider the same status for the statement s2=‘I am affected by disease D2’ and the
statement s1? [....]
Consider [as well the] statement s3=‘I am affected by D3’, where D3 is a poorly studied and initially asymptomatic disease you only know by name.'' \cite[p. 793--794]{baroni.2019}\\
\newline
The authors in \cite{baroni.2019} appeal to the intuition that there should be a different justification status for the statement s1 and the statement s2, although in both cases there are arguments for \textit{and} arguments against the statement. Moreover, we want to say that the status of statement s3 should be different from the statuses of s1 and s2, too. s3 should intuitively get a justification status of ``full ignorance'' as there is no evidence that concerns the third disease (and thereby statement s3).\\
In fact, the evaluation of the different labellings\footnote{They first investigate different formalism of structured argumentation and how the different types of labellings can be implemented there. They then investigate which outcomes the different labellings yield for the provided example.
For some formalisms they distinguish between a sceptical and a credulous approach. This distinction only yields a different treatment of statement s1 and its negation, but is silent on the other statements. For reason of simplicity, we will only consider the results of the sceptical labellings.} by means of the example shows that only the ignorance-aware labelling can account for this intuition. The different labellings provide the following results:\\

\begin{tabular}[t]{lccccc}
 & s1 & $\lnot$s1 & s2 & $\lnot$s2 & s3 \\
 
Bivalent Labellings  & $\mathrm{no}$ & $\mathrm{no}$ & $\mathrm{no}$ & $\mathrm{yes}$ & $\mathrm{no}$ \\

Doubt-Tolerant Labellings & $\mathrm{ni}$ & $\mathrm{ni}$ & $\mathrm{fal}$ & $\mathrm{yes}$ & $\mathrm{ni}$ \\

Ignorance-Aware Labellings & $\mathrm{ni}$ & $\mathrm{ni}$ & $\mathrm{fal}$ & $\mathrm{yes}$ & $\mathrm{unk}$ \\
\end{tabular}\\
\newline
One can see that the bivalent labellings cannot even illustrate the different intuitive justification statuses of statement s1 and s2. The doubt-tolerant labellings already do a better job, recognising that the question about s1 is, in comparison to s2, not decided, and hence taking both s1 and its negation to be $\mathrm{ni}$, while s2 is labelled $\mathrm{fal}$ and its negation is labelled $\mathrm{yes}$. However, there is no way to distinguish the justification status of s1 (and $\lnot$s1) from the justification status of s3. This is only achievable in the ignorance-aware labellings. Recall that ignorance-aware labellings distinguish between two forms of the middle, undetermined status: the label $\mathrm{ni}$ represents ``conflicting support'', while $\mathrm{unk}$ represents the ``absence of support'', \cite[p. 848]{baroni.2019}. This fits our intuitions of the example, as s3 is a statement, for which there is no support at all, while s1 is a statement for which the support is conflicting.\\
\newline
With the two different labels from ignorance-aware labellings that represent indecision, the authors in \cite{baroni.2019} want to distinguish cases where the evidence is absent from cases where the evidence is equally balanced. This distinction is exactly reflected in the two basic norms for indecision from epistemology: the absence of evidence norm and the balanced evidence norm.\\
Moreover, the different forms of indecision, found in our philosophical range are represented in the example and covered (at least to some extent) by the ignorance-aware labellings. 
As we have seen, there is no evidence at all, speaking for or against s3, so s3 is labelled $\mathrm{unk}$. The subject has not even considered s3, and hence it is comparable to the statement ``The guava fruit has a lot of vitamin C'', that I have never considered. Thus, in the range of indecision, s3 similarly has to be placed quite at the beginning of the axis of engaging. It is clear that we have some form of ``not engaged'' indecision, as the subject is not concerned with the statement at all. 
On the other hand, s1 is a statement the subject already collected evidence for. There is one physician arguing for s1 and another, equally competent physician, arguing against s1. The subject has engaged with the statement in question and came to the conclusion that she cannot tell, whether she has the disease. This is clearly a form of more engaged indecision, as the subject not only considered the statement s2, but actually collected quite some evidence for (and against) it. This can be compared to the example of me wondering whether the covid pandemic will be over next year. I, too, collected evidence for and against the proposition (and in fact new evidence gets through to me every day), but I still cannot decide, because the evidence seems more or less balanced.\\
\newline
We see that the ignorance-aware labellings allow us to distinguish two distinct forms of indecision. It might be valuable, however, to represent even more than two forms of indecision in this framework. When we look back to the axis that represented the example situations of \ref{example phil}, we already find four distinct forms of indecision. As said above, the situation of me being undecided about the Guava fruit's nutritional value can be compared to the indecision concerning statement s3, saying that I have disease D3 that I never heard of. Both cases represent a rather passive and not engaged form of indecision as I have not considered the respective statement or proposition. In the philosophical example, however, the cavemen's case exemplifies another form, that is even more left on the axis and hence even less engaged. In argumentation theory, it can also be eligible, to make a more fine-grained distinction between such little-engaged forms. Although s3 is a statement that has not been considered yet, the reasoning subject or system at least understands the statement. However, there can be cases in which the system cannot grasp or process the statement, when, for example, the statement contains words or phrases which do not belong to the language of the respective argumentation theory system. Such cases would correspond to the cavemen being undecided\footnote{As said above, philosophers would not use the term indecision, but mere non-belief here.} about whether quarks exist. In such a case, the system should be able to reply a different form of indecision than in the case of s3. While s3 is a statement with respect to which the system is not equipped with any argument for or against it, it still could in principle evaluate the statement with a decisive label if, for example, at some later point new arguments would come into play. With a statement that is not even included in the system's language, this is different. The system here should give a reply that shows full ignorance about what to to with that statement.\\
On the other side of the axis, there is room for finer distinctions, too. Philosophers distinguish between rather open, but yet reflective forms of indecision (the Covid case) from settled forms of indecision (the question about God's existence). As we have seen, the example of s1 and $\lnot$s1 represent a similar situation as the Covid case, as the subject will be undecided whether she has disease D1 but will still look for further evidence, possibly changing the label of the statement with the help of further arguments. There might be other examples, however, where the argumentation framework is build in a way in which it is clear that a statement \emph{cannot} get a different label than that of being undecided. The cycles of attack that have been regarded in section \ref{Forms of Indecision on the Argument Level} provide an outline of a possible example of such situations. To distinguish these two forms with two different labels can be helpful for a system to recognise when there is no need for further deliberation or further arguments concerning a certain statement, and when it is worth to allow further arguments in order to reevaluate. 

\section{Conclusion and Outlook}\label{Conclusion and Outlook}
In this paper I laid out some first steps toward a better understanding of the forms of indecision involved in argumentation theory. For this end, I presented philosophical considerations concerning different phenomena related to indecision and norms about when a subject should or should not be undecided about a matter. I tried to show that a lot of the philosophical investigations find some correspondence both on the argument level of abstract argumentation theory and on the level of statement evaluation. 
By having revealed parallels of this kind, we took the first steps toward using philosophical arguments to better evaluate how indecision is used in argumentation theory. The philosophical considerations can, for example, be used to evaluate the status of indecision in different semantics. Once we have drawn these parallels, and understood the use of indecision and the different forms involved in argumentation theory, we are able to apply further developments in philosophy concerning indecision or other neutral stances to argumentation theory, too. This is interesting, because indecision can be a useful tool for indicating unclear or critical situations.\\
\newline
For this enterprise, investigating indecision on a statement level seems crucial. For further research, it is, hence, desirable to concentrate on the different ways to evaluate statements and on the different ways to transfer justification of arguments to justification of statements, yielding statement-labellings. Although recent work already allows for a distinction between two kinds of indecision-labels on a statement level, I argued that it is necessary to distinguish even more forms of indecision. It is desirable that an argumentation system is capable of distinguishing cases in which it cannot deal with the input, from cases in which it has no evidence for or against a statement; and cases for which the evidential situation seems balanced, so it might need further arguments, from cases that clearly cannot be decided. Additionally, the system should be capable of storing the information and reporting these different situations. These skills can help the system to signal when it reaches its limits of application or when it needs further evidence or arguments to decide. This can help to reduce random decisions and make the systems choices more grounded and traceable.\\
Another challenge in the field of statement evaluation will be to not only consider the conclusions of an argument but also the premises. This is especially relevant when considering the different forms of defeat, such as undercutting or rebutting defeat, see \cite{pollock.1995}.\\
On the other hand, it can also be interesting to take a look at an even higher level than arguments. When non-unique semantics are considered, the question about which labelling (or extension) is to choose suggests itself immediately. For example, Dauphin et. al. investigate in \cite{dauphin.2018} certain principles that decision graphs that are used for choosing among multiple extensions should satisfy.
Concerning such choosing strategies, a better understanding of the underlying concepts of indecision can also be very useful for cases in which the available evidence does clearly point to one direction.

\bibliographystyle{deon16}
\bibliography{deon16}

\end{document}